\documentclass[
]{ceurart}

\usepackage{listings}
\usepackage{float}
\begin{document}

\copyrightyear{2025}
\copyrightclause{Copyright for this paper by its authors.
  Use permitted under Creative Commons License Attribution 4.0
  International (CC BY 4.0).}

\conference{IJCAI-25 Diversity and Inclusion Program: Empowering Women of Colour in AI-Driven Mental Health Research Workshop, August 18th, 2025, Montreal, Canada. This work is accepted and presented at the workshop. A revised version was invited for consideration in the CEUR
Workshop Proceedings and is currently under review.}

\title{Handling Extreme Class Imbalance: Using GANs in Data
Augmentation for Suicide Prediction}

\author[1]{Vaishnavi Visweswaraiah}[orcid=0009-0007-0883-6899,email=vvisweswaraiah@my.harrisburgu.edu,]
\cormark[1]
\fnmark[1]
\address[1]{Information Systems Engineering and Management (Harrisburg University of Science \& Technology), Harrisburg, USA}

\author[2]{Tanvi Banerjee}[orcid=0000-0002-9794-3755,email=tanvi.banerjee@wright.edu,]
\fnmark[1]
\address[2]{Computer Science and Engineering (Wright State University), Fairborn, USA}

\author[3]{William Romine}[orcid=0000-0002-0386-1688,email=william.romine@wright.edu,]
\fnmark[1]
\address[3]{Biological Science (Wright State University), Fairborn, USA}

\cortext[1]{Corresponding author.}
\fntext[1]{These authors contributed equally.}
\begin{abstract}
  Suicide prediction is the key for prevention, but real data with sufficient positive samples is rare and causes extreme class imbalance. We utilized machine learning (ML) to build the model and deep learning (DL) techniques, like Generative Adversarial Networks (GAN), to generate synthetic data samples to enhance the dataset. The initial dataset contained 656 samples, with only four positive cases, prompting the need for data augmentation. A variety of machine learning models, ranging from interpretable data models to black box algorithmic models, were used. On real test data, Logistic Regression (LR) achieved a weighted precision of 0.99, a weighted recall of 0.85, and a weighted F1 score of 0.91; Random Forest (RF) showed 0.98, 0.99, and 0.99, respectively; and Support Vector Machine (SVM) achieved 0.99, 0.76, and 0.86. LR and SVM correctly identified one suicide attempt case (sensitivity: 1.0) and misclassified LR(20) and SVM (31) non-attempts as attempts (specificity: 0.85 \& 0.76, respectively). RF identified 0 suicide attempt cases (sensitivity: 0.0) with 0 false positives (specificity: 1.0). These results highlight the models' effectiveness, with GAN playing a key role in generating synthetic data to support suicide prevention modeling efforts.
\end{abstract}


\maketitle
\section{Introduction}
Suicide is a leading cause of death in the United States (U.S) and worldwide. In the U.S., suicide rates increased by approximately 36\% between 2000 and 2022 \cite{cdc_facts_2025,stone_notes_2023}. In 2023, suicide was the second leading cause of death for people aged 10–34, which includes adolescents and young adults \cite{cdc_facts_2025}. Early detection and intervention are critical but often inadequate due to reliance on self-reporting. Machine learning (ML) and Deep Learning (DL) techniques can be used in suicide prediction by analyzing patients’ clinical and psychological factors. However, limited data continues to be a major challenge for sensitive topics such as suicide.

Through data augmentation, we aim to improve the performance of ML models. The paper details the methodology where the entire process, including data splitting, preprocessing, grid search, and Generative Adversarial Networks (GAN) based augmentation, is visually summarized in appendix~\ref{appendix:overall_experiment_architecture} Fig.~\ref{fig:experiment_flow}.

This overarching methodology and its outcomes form the core idea of this study. Additionally, a comparative analysis of model performance is presented to draw conclusions about which models are most effective in detecting rare suicide attempts and how GAN-based data augmentation influences their behavior, contributing to the role of ML in suicide prevention and laying the groundwork for future research. This work addresses the following research questions: (1) Which ML models best identify suicide attempt cases under extreme class imbalance? and (2) Can synthetic data generated via GANs improve the detection of rare, high-risk cases?

\section{Related Work}

Active research has focused on applying ML techniques to develop tools in predicting suicide with more accuracy and efficiency. Prior studies have used ML models, such as neural networks, to predict suicidal ideation and adolescent suicide attempts based on clinical data, comparing configurations with traditional methods \cite{ji_suicidal_2021,bhat_predicting_2017}. GANs have become a key approach for data augmentation, generating synthetic datasets like time-series \cite{smith_conditional_2020}, tabular \cite{xu_modeling_2019}, and medical imaging data \cite{dimitrakopoulos_ising-gan_2020,waheed_covidgan_2020}, improving classification performance and addressing data scarcity. GANs have also been applied to one-dimensional medical data \cite{zhang_gan-based_2023} and sensor-based health data using Long Short-Term memory (LSTM) based time-series GANs for medical diagnosis \cite{yang_ts-gan_2023}. Applications extend to augmenting Internet of Medical Things (IoMT) data, including chronic obstructive pulmonary disease monitoring \cite{vaccari_generative_2021}, skin cancer detection \cite{rashid_skin_2019}, and drug usage prediction using high-dimensional, low-sample-size data \cite{tanaka_data_2019,thach_novel_2024}. Despite their promising results, obtaining sufficient and balanced behavioral health data remains challenging for suicide prediction. Our research leverages GAN methods to generate synthetic data, addressing imbalance and scarcity to build reliable tools for suicide prevention.

\section{Experiment}
\subsection{Initial and GAN Data Exploration}
The dataset used in this experiment is taken from the Adolescent Brain Cognitive Development (ABCD) Study \cite{noauthor_abcd_nodate}. Rasch modeling was used to translate raw survey question responses into the unidimensional outcomes of Panic and Suicidal Ideation. Rasch models are standard practice for the formulation of objective measures from test and survey responses \cite{Wright1979Best}. Unidimensional log-odds (“logit”) measures were calculated for each outcome. Panic was calculated from the reporting of seven symptoms (within the last 2 weeks): (1) shortness of breath, (2) shaking, (3) palpitation, (4) nausea, (5) sweating, (6) dizziness, and (7) choking.  These were reported dichotomously and generated standardized measures for panic with an internal consistency (Cronbach’s alpha) of 0.60. Suicidal Ideation was calculated using the Rasch model from the reporting of four questions:
\begin{itemize}
 \item (1)   Sometimes when kids get upset or feel numb, they may do things to hurt themselves, like scratching, cutting, or burning themselves. In the past two weeks, how often have you done any of these things or other things to try to hurt yourself? (Ordinal 1-4)
\item (2)   Was there ever a time in the past when you did things to hurt yourself on purpose because you were upset, like cut, scratch or burn yourself? (Dichotomous)
\item (3)   In the past two weeks, how often have you wished you were dead or had thoughts that you would be better off dead? (Ordinal 1-4)
\item (4)   In the past two weeks, how often did you think seriously about wanting to kill yourself? (Ordinal 1-4)
\end{itemize}

Given the mix of dichotomous and ordinal scoring, a Rasch partial credit  \cite{Masters_1982} was used to generate the logit measure for Suicidal Ideation with an internal consistency (Cronbach's alpha) of 0.58.  For more details on data source and code availability, please refer to (appendix~\ref{appendix:dataset_details}).

The initial dataset contains 656 samples classified into two categories based on suicide attempts: 652 negative samples (label 0) and four positive samples (label 1), with features including sex, panic, and suicidal ideation. After pre-processing data, the final dataset contained 655 samples with 651 negative and four positive samples. Due to this extreme class imbalance, we used Conditional GAN (appendix~\ref{appendix:gan_datageneration}) to generate synthetic samples, which is 40\% of the original dataset size, resulting in a balanced dataset with 134 negative samples and 126 positive samples. 

\begin{figure}[t]
    \centering
    \begin{minipage}{0.45\textwidth}
        \centering
        \includegraphics[width=\linewidth]{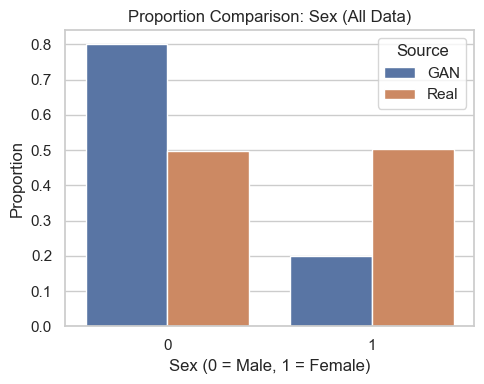}
        \par\vspace{0.5em}
        \textbf{(a)} Feature: sex – 0 = male, 1 = female
        \label{fig:s_sex_gan}
    \end{minipage}
    \hfill
    \begin{minipage}{0.45\textwidth}
        \centering
        \includegraphics[width=\linewidth]{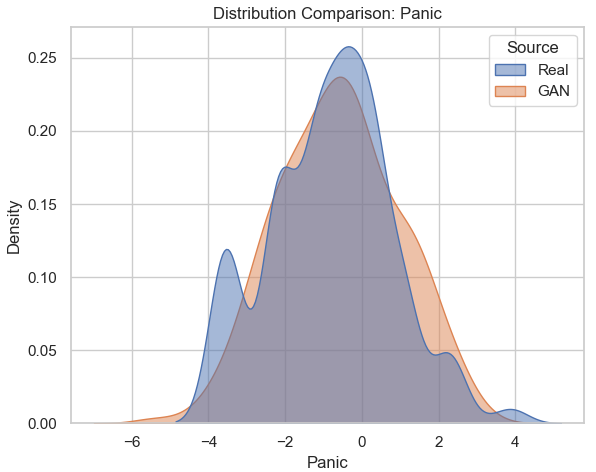}
        \par\vspace{0.5em}
        \textbf{(b)} Feature: Panic
        \label{fig:s_panic_gan}
    \end{minipage}

    \vspace{1em}
    \begin{minipage}{0.45\textwidth}
        \centering
        \includegraphics[width=\linewidth]{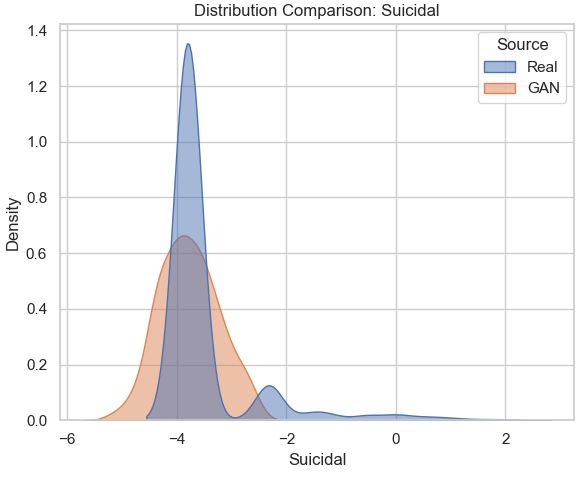}
        \par\vspace{0.5em}
        \textbf{(c)} Feature: Suicidal
        \label{fig:s_suicidal_gan}
    \end{minipage}

    \caption{ real vs gan data distribution for all samples}
    \label{fig:real_gan_data_analysis}
\end{figure}

To validate the GAN-generated data, we compared the distribution of key features (sex, panic, and suicidal) across both real and GAN data. The generated data showed alignment with real data w.r.t panic and suicidal features. For instance, the panic feature distribution for both real and GAN peaks at almost the same place and also has similar central tendencies, though the GAN distribution curve is a bit smoother(Fig.~\ref{fig:real_gan_data_analysis}(b)). Similarly, the distribution of Suicidal features for GAN peaks around the same low-suicidal region as real data, though the GAN curve is slightly wider; it did replicate the real suicidal distributions (Fig.~\ref{fig:real_gan_data_analysis}(c)). However, GAN fails to match the overall proportional balance for the 'sex' feature. In real data, males and females are equally proportioned (Fig.~\ref{fig:real_gan_data_analysis} (a) (Orange)), but GAN data generated an imbalance, resulting in a higher proportion of males compared to females (Fig.~\ref{fig:real_gan_data_analysis}(a) (Blue)). To analyze this feature further, we compared the proportions for positive suicide-attempt label data. This analysis resulted in the positive suicide attempt group (Fig.~\ref{fig:gan_data_exploration} (a) (Red)) having a higher proportion of females than the non-attempt group (Fig.~\ref{fig:gan_data_exploration} (a) (Blue)), indicating that though gan introduced overall male bias, it did learn females are more likely to attempt suicide, reflecting the real-data pattern of suicide attempt distribution (Fig.~\ref{fig:real_data_exploration} (a) (Red)).
We further analyzed the real and GAN-generated data individually at the label level to get deeper insights.

\subsubsection{Real Data}
\begin{figure}[h!]
    \centering
    \begin{minipage}{\textwidth}
        \centering
        \includegraphics[width=0.6\linewidth]{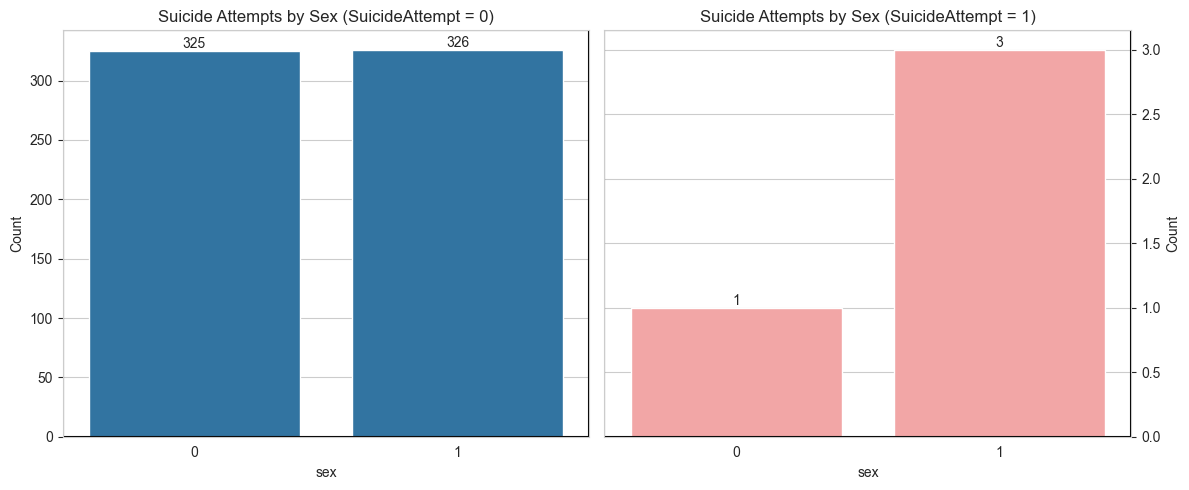}
        \par\vspace{0.5em}
        \textbf{(a)} Feature: sex – 0 = male, 1 = female
        \label{fig:s_sex}
    \end{minipage}
    \hfill
    \begin{minipage}{\textwidth}
        \centering
        \includegraphics[width=0.6\linewidth]{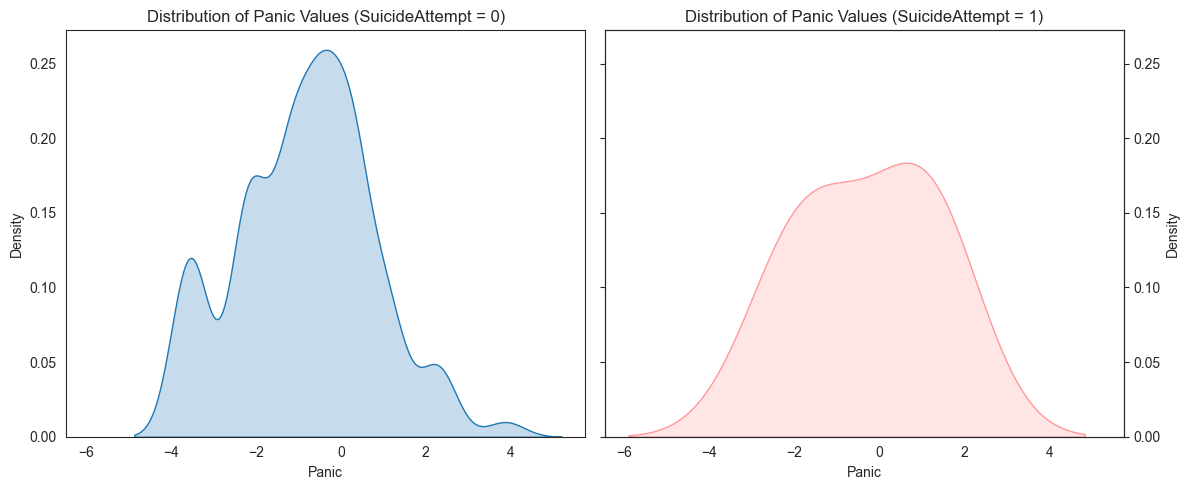}
        \par\vspace{0.5em}
        \textbf{(b)} Feature: Panic
        \label{fig:s_panic}
    \end{minipage}

    \vspace{1em}
    \begin{minipage}{\textwidth}
        \centering
        \includegraphics[width=0.6\linewidth]{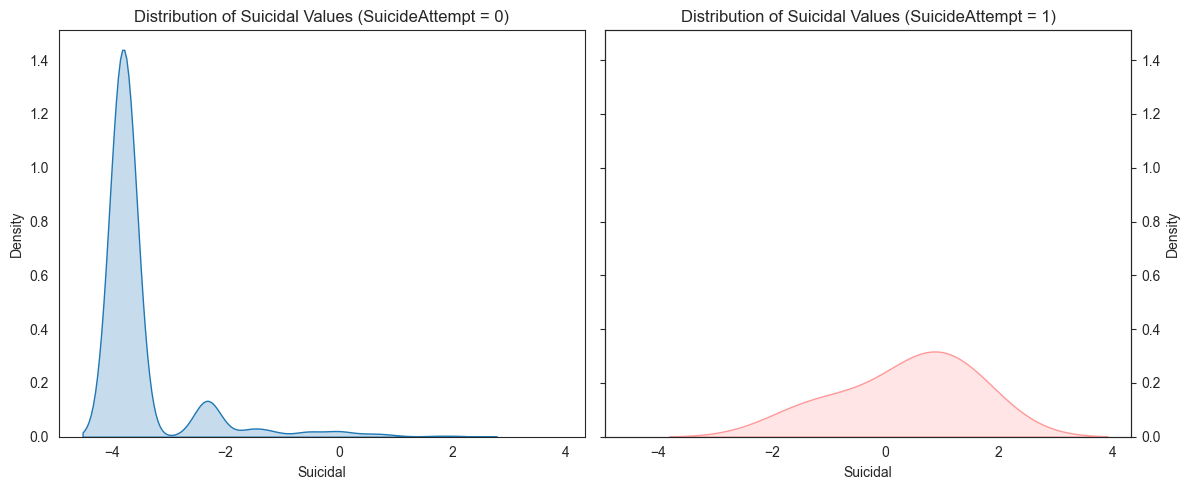}
        \par\vspace{0.5em}
        \textbf{(c)} Feature: Suicidal
        \label{fig:s_suicidal}
    \end{minipage}

    \caption{Real Data distribution for all suicide attempt samples ( label-0 : no suicide attempt (Blue), label-1: suicide attempt (Red)}
    \label{fig:real_data_exploration}
\end{figure}

The real data set contains 655 samples, comprising 652 negative and four positive samples. For negative samples (no suicidal attempt), 
Fig.~\ref{fig:real_data_exploration}(a) (Blue), Fig.~\ref{fig:real_data_exploration}(b) (Blue), and Fig.~\ref{fig:real_data_exploration}(c) (Blue) show (a) an equal distribution of males and females, suggesting gender does not play a significant role, (b) a irregular bell-shaped curve for panic scores centered around lower values, and (c) suicidal levels peaking at lower scores, respectively. For positive samples, Fig.~\ref{fig:real_data_exploration}(a) (Red), Fig.~\ref{fig:real_data_exploration}(b) (Red), and Fig.~\ref{fig:real_data_exploration}(c) (Red) reveal given insights:(a) a higher number of suicide attempts among females compared to males (3:1 ratio), (b) panic values shifting towards higher values, and (c) suicidal values increasing, indicating a strong correlation with suicide attempts. These patterns align with clinical expectations and emphasize the importance of these features for prediction.

\subsubsection{Synthetic GAN Data}
\begin{figure}[h!]
    \centering
    \begin{minipage}{\textwidth}
        \centering
        \includegraphics[width=0.6\linewidth]{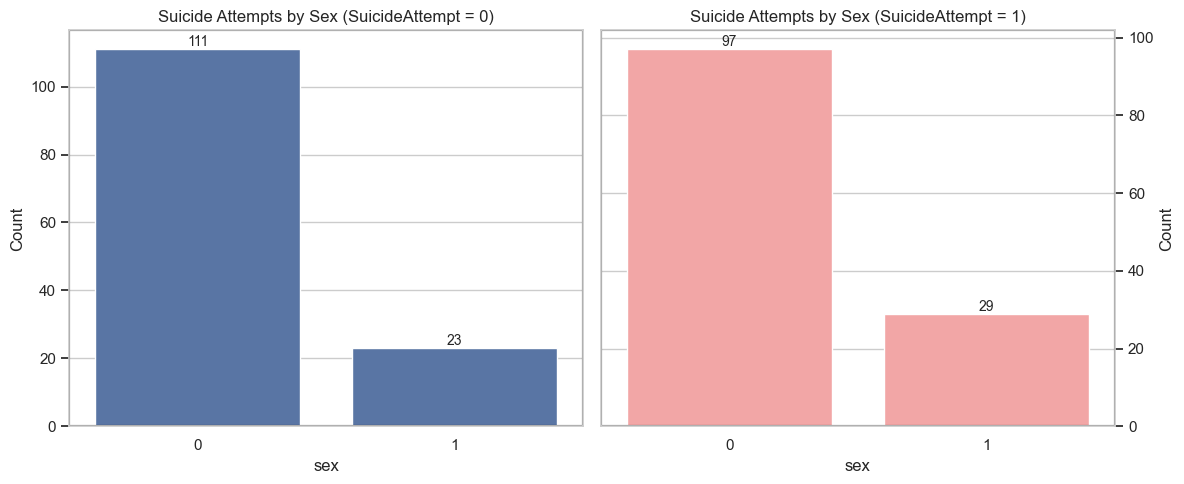}
        \par\vspace{0.5em}
        \textbf{(a)} Feature: sex – 0 = male, 1 = female
        \label{fig:ns_sex_gan}
    \end{minipage}
    \hfill
    \begin{minipage}{\textwidth}
        \centering
        \includegraphics[width=0.6\linewidth]{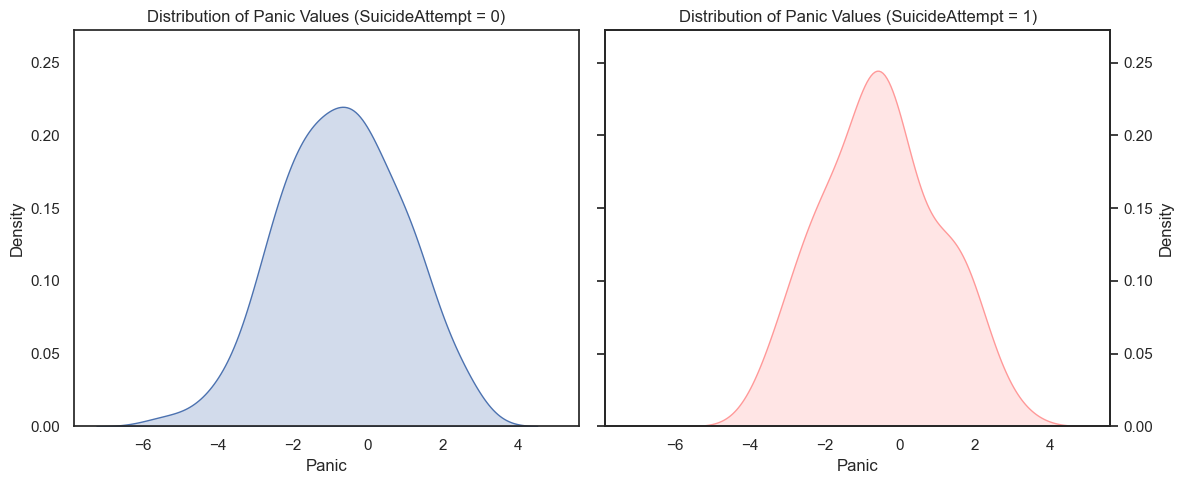}
        \par\vspace{0.5em}
        \textbf{(b)} Feature: Panic Scores
        \label{fig:ns_panic_gan}
    \end{minipage}

    \vspace{1em}
    \begin{minipage}{\textwidth}
        \centering
        \includegraphics[width=0.6\linewidth]{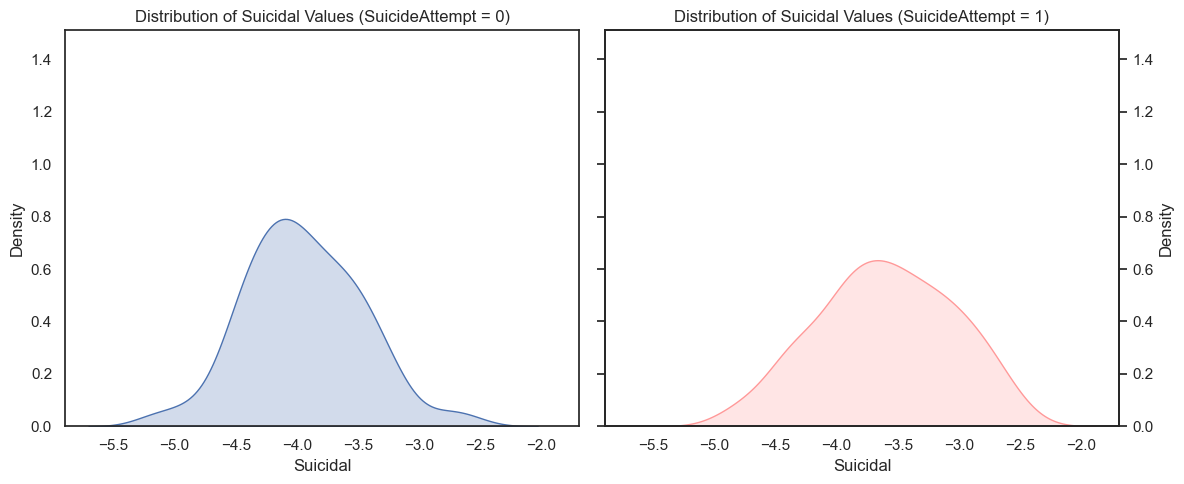}
        \par\vspace{0.5em}
        \textbf{(c)} Feature: Suicidal Scores
        \label{fig:ns_suicidal_gan}
    \end{minipage}

    \caption{GAN Data distribution for all suicide attempt samples ( label-0 : no suicide attempt (Blue), label-1: suicide attempt (Red)}
    \label{fig:gan_data_exploration}
\end{figure}

GAN-generated data containing 40\% of the original dataset size contains 134 negative samples and 126 positive samples, whose data exploration for features like panic values and suicidal values differs between the two classes. For the non-attempt group, Fig.~\ref{fig:gan_data_exploration}(a) (Blue), Fig.~\ref{fig:gan_data_exploration}(b) (Blue), and Fig.~\ref{fig:gan_data_exploration}(c) (Blue) show (a) a higher proportion of males than females, (b) panic values are normally distributed, peaking at moderate values, and (c) suicidal values peak toward lower scores. For the suicide attempt group, Fig.~\ref{fig:gan_data_exploration}(a) (Red), Fig.~\ref{fig:gan_data_exploration}(b) (Red), and Fig.~\ref{fig:gan_data_exploration}(c) (Red) reveal (a) more suicide attempts among males than females, but with a relatively higher proportion of females compared to the non-attempt group Fig.~\ref{fig:gan_data_exploration}(a) (Blue), (b) panic scores shifting slightly toward higher values, and (c) suicidal values peaking at higher scores compared to the non-attempt group. Fig.~\ref{fig:gan_data_exploration} shows apparent differences in demographics and mental health indicators between the two groups. While males dominate both groups, the suicide attempt group shows a slightly higher proportion of females. On the other side, panic and suicidal levels are elevated in the suicide attempt group, emphasizing the risk factors.

\subsection{Model Training \& Results }
We trained three ML models, LR, SVM, and RF, on two datasets: (1) a real train set (521 negative, 3 positive) with leave-one-out cross-validation F1 scores (0.006 for LR and SVM, 0.000 for RF) despite high accuracy (0.884 for LR, 0.994 for SVM, 0.781 for RF), and (2) GAN-generated data (134 negative, 126 positive) with cross-validation F1 scores (0.312 for LR and 0.285 for SVM, 0.304 for RF), and moderate accuracy (0.642 for LR, 0.608 for SVM, 0.612 for RF (Appendix~\ref{appendix:training}).

As detailed in Appendix~\ref{appendix:results}, performance was evaluated on a real-world test set comprising 130 negative and 1 positive sample. In this evaluation, both LR and SVM—trained on original data—successfully identified the sole positive case, with LR producing 20 false positives and SVM 31. RF, however, failed to detect the positive instance. In contrast, all GAN-trained models (LR\_G, SVM\_G, RF\_G) were able to identify the single positive case, with LR\_G and RF\_G each producing a high number of false positives (87 and 85, respectively), and SVM\_G producing 33 (Table~\ref{tab:test_confusion_matrix}). \textit{This demonstrates that training on synthetic data can introduce signals helpful for rare-case detection, even for models like RF that fail under extreme imbalance.}

Among all models trained on GAN data, SVM\_G showed the most balanced performance: it maintained the ability to detect the positive case (recall improved from 0 to 1) while keeping false positives (33) considerably lower than LR\_G and RF\_G. This suggests that SVM not only generalized well when trained on limited real data but also adapted effectively when trained entirely on synthetic data. Its precision-recall trade-off remained stable across training sources, making SVM a promising choice in scenarios where missing positive cases carries high cost.

\section{Discussion} 

We have developed a process to generate and evaluate the model's performance on synthetic data generated using a conditional GAN, which is important when dealing with severe class imbalance. Conditional GAN allows us to generate the minority (positive) class in suicide risk prediction, unlike its more popular counterpart, Wasserstein GAN (WGAN). Although WGANs are well known for more stable training and reducing model collapses, \textit{they do not generate class-specific data unless modified to conditional WGAN} \cite{engelmann2020conditionalwassersteinganbasedoversampling}.

Our findings highlight that simpler models like LR and SVM are better suited for generalizing under conditions of severe data imbalance. Notably, both LR and SVM models trained on original data successfully identified the positive suicide case in the real test set, whereas RF failed to do so. However, when trained on GAN-generated data, even RF (RF\_G) was able to identify the single positive test case, demonstrating that GAN-based data augmentation, even though with a male-skewed distribution (Fig.~\ref{fig:real_gan_data_analysis} (a) ) can still inject a useful signal that tree-based models might miss. This gain in sensitivity, however, came at the cost of specificity, as seen in the increased number of false positives(Table~\ref{tab:test_confusion_matrix}).

Among all models, SVM exhibited the most balanced and consistent performance. When trained on real data, SVM demonstrated a strong macro recall of 0.88 with moderate false positives (31). More importantly, when trained on GAN-generated data (SVM\_G), it retained this balance—successfully detecting the rare positive case while keeping false positives (33) lower than other GAN-trained models like LR\_G (87) and RF\_G (85), despite the 'sex' feature imbalance in the synthetic GAN dataset. This indicates that SVM generalized well not just across real-world data, but also across synthetic data, making it a particularly robust option for imbalanced clinical datasets.

One likely reason for SVM's consistent performance is its margin-based optimization approach, which allows it to construct decision boundaries that remain stable even when trained on synthetic patterns. Unlike RF, which is sensitive to noise in high-dimensional or class-skewed settings due to its tree-splitting mechanisms, or LR, which tends to be conservative in detecting rare cases, SVM though its false positive rate was higher than LR, it remained substantially lower than LR\_G and RF\_G, indicating SVM’s capacity to generalize under both real and synthetic training conditions. In the context of suicide prevention, this consistent recall with controlled false alarms is clinically meaningful: flagging a non-suicidal individual may result in a false alert, but missing a true suicidal case could have life-threatening consequences.

Performance of a machine learning model should always be interpreted in the context of the application domain. In suicide prediction, missing even a single positive sample (false negative) can be far more harmful than triggering a false alarm (false positive). These findings reinforce the need for thoughtful model selection, context-aware evaluation, and deeper scrutiny of how synthetic data affects predictions—especially in high-stakes, imbalanced healthcare scenarios.

GAN-generated data, when properly validated, holds promise for supporting rare-event detection such as suicide risk prediction but also more broadly in rare health-related event AI applications. It also emphasizes the importance of moving beyond traditional evaluation metrics to reflect the cost of both false positives and false negatives in sensitive domains.

In summary, our results show that when real data is limited, synthetic data—particularly when generated by class-conditional GANs—can support training of models for rare, high-impact predictions. SVM emerged as the most generalizable model across both original and synthetic conditions, highlighting the role of tailored data augmentation under extreme class imbalance.

\section{Limitations and Future Work}

Key challenges in this study include issues of high false positives in GAN-trained models and the failure of certain models, like Random Forest, to detect rare cases when trained on original data. While GAN effectively mimicked real data distributions, it failed to preserve the relationship of one feature ('sex' feature) by generating male-skewed data indicating a need for further refinement. Importantly, the evaluation of synthetic data lacked clinical validation. 

Future research can expand on this work by applying alternative generative techniques such as variational autoencoders (VAEs), or leveraging pre-trained transformer models to synthesize contextual behavioral data through prompt engineering. Cost-sensitive learning, threshold tuning, and clinician-in-the-loop evaluations, such as involving consultation with mental health professionals, will be crucial to improving the clinical relevance and deployment readiness of such systems.

\section*{Declaration on Generative AI}
  
 During the preparation of this work, the author(s) used Grammarly to check grammar and spelling. After using these tool(s)/service(s), the author(s) reviewed and edited the content as needed and take full responsibility for the publication’s content. 

\bibliography{SuicidePrediction}

\appendix
\section{Overall Experiment Architecture}
\label{appendix:overall_experiment_architecture}
\begin{figure}[htbp!]
    \centering
    \includegraphics[width=0.9\linewidth]{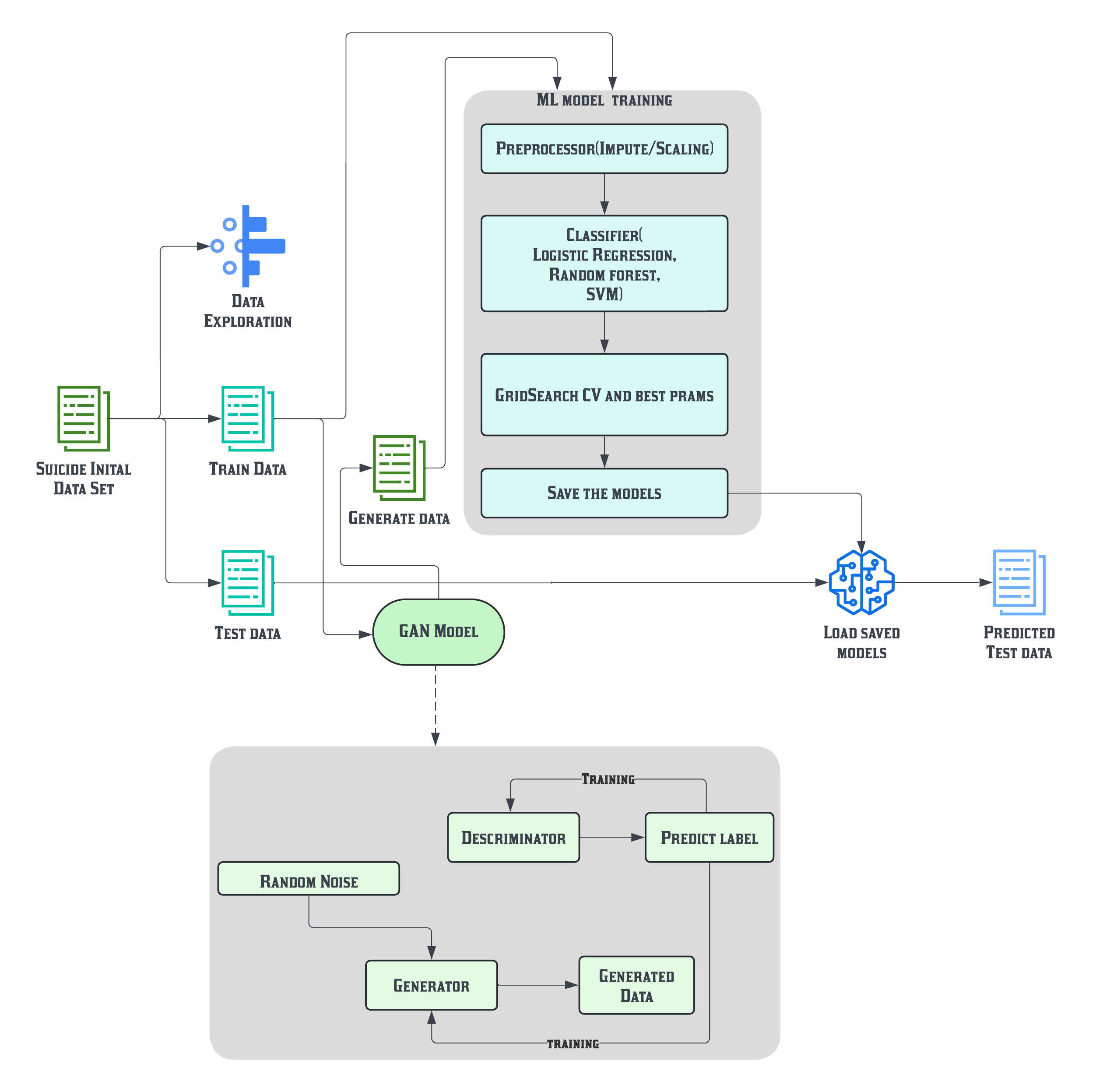}
    \caption{Overall Experiment Architecture}
    \label{fig:experiment_flow}
\end{figure}

\section{Dataset}
\label{appendix:dataset_details}
The dataset used in this experiment is taken from the Adolescent Brain Cognitive Development (ABCD) Study, a longitudinal, multisite project tracking over 10,000 children aged 9-10 into early adulthood, focusing on cognitive, social, emotional, and physical development.
\paragraph*{Data and Code Availability}
Data and code are not publicly available. Data used in the preparation of this article were obtained from the Adolescent Brain Cognitive Development (ABCD) Study \footnote{https://abcdstudy.org}, held in the NIMH Data Archive (NDA). This is a multisite, longitudinal study designed to recruit more than 10,000 children age 9-10 and follow them over adolescence into young adulthood. The ABCD Study® is supported by the National Institutes of Health and additional federal partners under multiple grant programs. A full list of supporters is available at \url{https://abcdstudy.org/federal-partners.html}. A listing of participating sites and a complete listing of the study investigators can be found at \url{https://abcdstudy.org/consortium\_members/}.
ABCD consortium investigators designed and implemented the study and/or provided data but did not necessarily participate in the analysis or writing of this report. This manuscript reflects the views of the authors and may not reflect the opinions or views of the NIH or ABCD consortium investigators. 
The (NIMH) Data Archive(NDA) required approval for data access\footnote{https://nda.nih.gov/abcd/request-access}\footnote{https://nda.nih.gov/ndapublicweb/Documents/NDA+Submission+Request.pdf} and approved data use certification. Hence, data is not available for distribution as part of the data use terms and conditions.

\paragraph*{Institutional Review Board (IRB)}
The National Institute of Mental Health (NIMH) Data Archive (NDA) is an NIH-funded collaborative resource that
contains harmonized human subjects research data and metadata from multiple research Data Repositories, providing a rare and valuable scientific resource. Data submitted to NDA have been stripped of all individual identifiers. The data accessed and used in this research does not contain any PII.

\section{Experiments}
\label{appendix:ml_experiment}

This paper discusses a methodology or process of generating augmented data and how to evaluate it; this paper uses novel ML and deep learning algorithms like GAN for data generation and model training on three ML algorithms: Logistic Regression, SVM, and Random Forest. It later evaluates the results on real test data.

\subsection{Data Generation Using GAN}
\label{appendix:gan_datageneration}

\begin{figure}[t]
    \centering
    \includegraphics[width=0.8\linewidth]{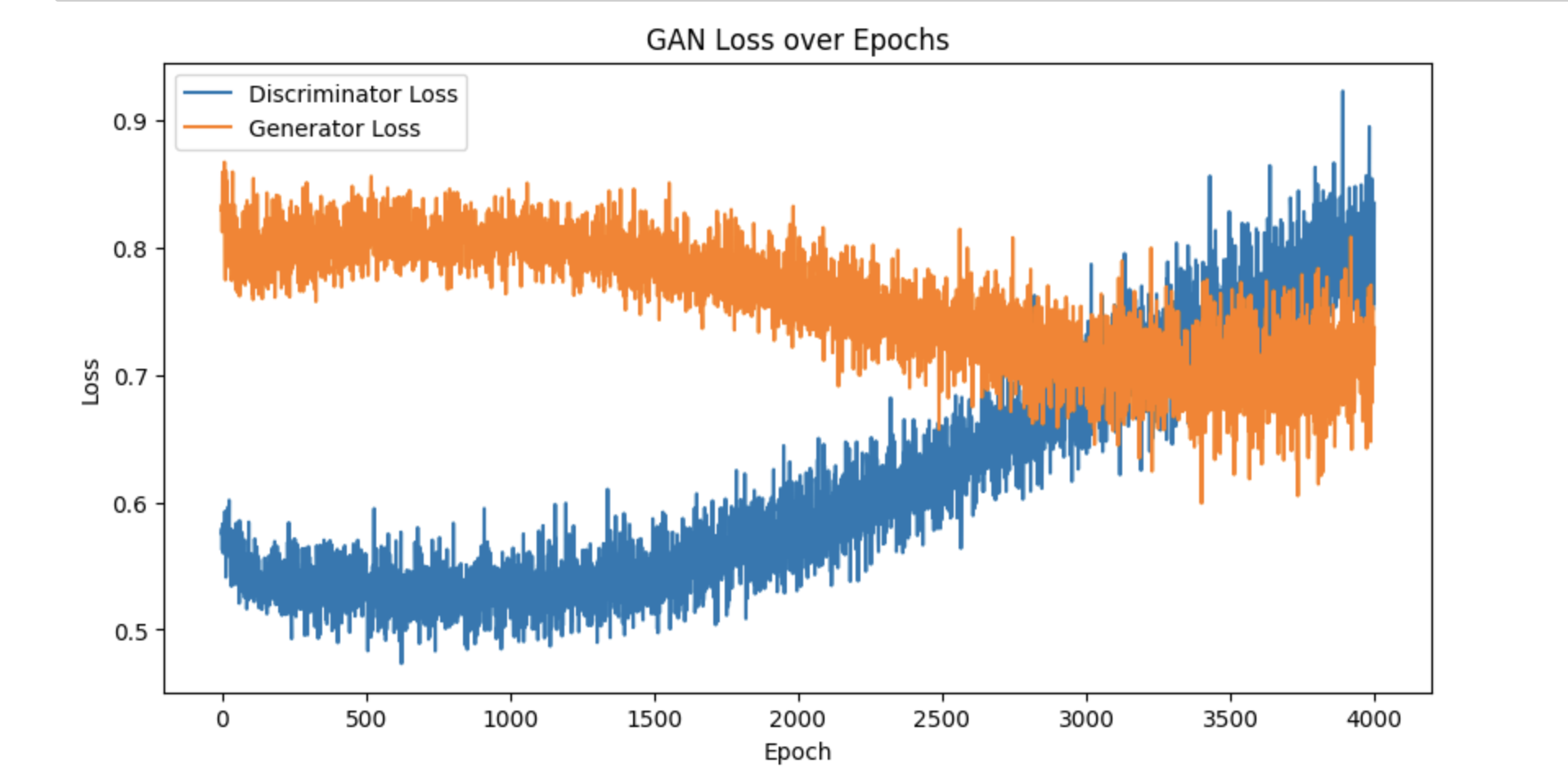}
    \caption{GAN loss highlighting the changes in generator loss versus discriminator loss over epochs}
    \label{fig:gan_epoch}
\end{figure}

A Conditional GAN extends the standard GAN by generating data conditioned on an additional input, such as a label or specific attribute\cite{xu_modeling_2019}. In our implementation, the generator consists of dense layers with activation functions and batch normalization, producing continuous data (e.g., 'Panic' and 'Suicidal' levels) using linear activation and categorical data (e.g., 'sex' and 'Female' derived from 'sex') using sigmoid activation. The discriminator distinguishes between real and synthetic data using dense layers with leaky ReLU activation and dropout layers to prevent overfitting, with a single sigmoid-activated output. The training involved multiple heuristic experiments, finalizing a learning rate of 0.00002 for the generator and 0.000025 for the discriminator. With the Adam optimizer and binary cross-entropy loss function, the discriminator loss reached 0.8, indicating the generator adequately challenges the discriminator while still being distinguishable. The generator loss was 0.7, reflecting effective progress in creating realistic data. 
The closeness of these loss values, as shown in Fig.~\ref{fig:gan_epoch}, suggests a balanced model, which is essential for generating high-quality synthetic data. This closeness of value suggests that neither the generator nor the discriminator dominates. This is a key factor in sensitive applications like behavioral health modeling, where the accurate simulation of complex human behaviors is essential.

\subsection{Model Training}
\label{appendix:training}
The original data is split into an 80:20 ratio using a stratification strategy to preserve the percentage samples of the class distribution. The training set consists of 524 samples (521 negative and three positive) and GAN-generated(synthetic) data ( 134 negative and 126 positive), while the test set contains 131 samples (130 negative and one positive). 

The dataset undergoes several pre-processing steps like handling null values with imputation, as well as normalizing features. Three machine learning algorithms—Logistic Regression (LR), Random Forest (RF), and Support Vector Machine (SVM)—are trained using a pipeline that integrates preprocessing, grid search for hyperparameter tuning, and Leave-One-Out (LOO) cross-validation.LOO cross-validation evaluates each data point as a validation set, ensuring maximum training data usage and comprehensive model assessment using metrics like F1 score, accuracy, precision, and recall. Based on heuristic experiments on these models, given outputs are derived. (1) Logistic Regression (LR): Serves as the baseline, achieving an F1 score of 0.006 on original data and 0.312 on synthetic data, with model coefficients in Table~\ref{tab:model_coefficients} and the evaluation metrics in Table~\ref{tab:model_cv_scores} summarize model performance.(2) Random Forest (RF): Handles imbalanced datasets but achieves an F1 score of 0.000 on original data and 0.285 on synthetic data as shown in Table~\ref{tab:model_cv_scores}.  
(3) Support Vector Machine (SVM): Performs well with scaled features, obtaining an F1 score of 0.006 on original data and 0.304 on synthetic data, with metrics also reported in Table~\ref{tab:model_cv_scores}.

\begin{table}[t]
\centering
\caption{Model coefficients of LR and RF trained on original and GAN-generated data}
\label{tab:model_coefficients}
\setlength{\tabcolsep}{4pt}
\begin{tabular}{lcccc}
\toprule
\textbf{Feature} & \textbf{LR} & \textbf{RF} & \textbf{LR\_G} & \textbf{RF\_G} \\
\midrule
Panic     & 0.10  & 0.20  & 0.03 & 0.40 \\
Suicidal  & 0.60  & 0.70  & 0.30 & 0.60 \\
Sex       & -0.10 & 0.10  & 0.05 & 0.02 \\
Female    & -0.10 & 0.01  & 0.05 & 0.02 \\
\bottomrule
\end{tabular}
\end{table}

\begin{table}[t]
\centering
\caption{Cross-validation scores of ML models on original and GAN-generated data}
\label{tab:model_cv_scores}
\setlength{\tabcolsep}{3pt}
\begin{tabular}{llcccc}
\toprule
\textbf{Model} & \textbf{Split} & \textbf{F1} & \textbf{Accuracy} & \textbf{Precision} & \textbf{Recall} \\
\midrule
LR     & Train & 0.090 & 0.884 & 0.047 & 1.000 \\
       & Val   & \textbf{0.006} & 0.884 & 0.006 & 0.006 \\
RF     & Train & 0.989 & 1.000 & 1.000 & 0.983 \\
       & Val   & \textbf{0.000} & 0.994 & 0.000 & 0.000 \\
SVM    & Train & 0.050 & 0.781 & 0.025 & 1.000 \\
       & Val   & \textbf{0.006} & 0.781 & 0.006 & 0.006 \\
LR\_G  & Train & 0.639 & 0.647 & 0.633 & 0.644 \\
       & Val   & \textbf{0.312} & 0.642 & 0.312 & 0.312 \\
RF\_G  & Train & 0.094 & 0.908 & 0.918 & 0.892 \\
       & Val   & \textbf{0.285} & 0.612 & 0.285 & 0.285 \\
SVM\_G & Train & 0.572 & 0.581 & 0.567 & 0.577 \\
       & Val   & \textbf{0.304} & 0.608 & 0.304 & 0.304 \\
\bottomrule
\end{tabular}
\end{table}

\section{Results}
\label{appendix:results}
 After training, all models, including model variations trained on synthetic data, are evaluated on the test data to determine their ability to predict positive suicide cases. As the original dataset is highly imbalanced, evaluation using micro and macro precision, recall, F1 scores, and weighted averages is required. These metrics ensure detailed performance assessment across both majority and minority classes, helping identify the rare yet crucial positive instances of suicide attempts. This detailed level of evaluation metrics is highly needed, as accurate predictions can save lives and every prediction carries huge consequences.
 
The original test data distribution has 130 negative cases and one positive case \{negative:130,positive:1\}. Despite being trained on only three positive samples, the model’s ability to predict a single positive case highlights its capabilities. Table~\ref{tab:model_test_scores} shows the test scores, and Table~\ref{tab:test_confusion_matrix} provides the confusion matrix for LR, RF, SVM, LR\_G, RF\_G and SVM\_G.

\begin{table}[t]
\centering
\caption{Test Scores for LR, RF, and SVM (Original and GAN-Trained)- *{\footnotesize Prec: Precision, Rec: Recall}}
\label{tab:model_test_scores}
\setlength{\tabcolsep}{4pt}
\begin{tabular}{llcccccc}
\toprule
\textbf{Model} & \textbf{Metric} & \textbf{Macro} & \textbf{Micro} & \textbf{Weighted} & \textbf{Label 0} & \textbf{Label 1} \\
\midrule
\multirow{3}{*}{LR} 
  & Prec & 0.52 & 0.85 & 0.99 & 1.00 & 0.05 \\
  & Rec  & \textbf{0.92} & \textbf{0.85} & \textbf{0.85} & \textbf{0.85} & \textbf{1.00} \\
  & F1   & 0.50 & 0.85 & 0.91 & 0.92 & 0.09 \\
\midrule
\multirow{3}{*}{RF} 
  & Prec & 0.50 & 0.99 & 0.98 & 0.99 & 0.00 \\
  & Rec  & 0.50 & 0.99 & 0.99 & 1.00 & 0.00 \\
  & F1   & 0.50 & 0.99 & 0.99 & 0.99 & 0.00 \\
\midrule
\multirow{3}{*}{SVM} 
  & Prec & 0.52 & 0.76 & 0.99 & 1.00 & 0.03 \\
  & Rec  & 0.88 & 0.76 & 0.76 & 0.76 & 1.00 \\
  & F1   & 0.46 & 0.76 & 0.86 & 0.86 & 0.06 \\
\midrule
\multirow{3}{*}{LR\_G} 
  & Prec & 0.50 & 0.34 & 0.99 & 1.00 & 0.01 \\
  & Rec  & 0.67 & 0.34 & 0.34 & 0.33 & 1.00 \\
  & F1   & 0.26 & 0.34 & 0.49 & 0.50 & 0.02 \\
\midrule
\multirow{3}{*}{RF\_G} 
  & Prec & 0.51 & 0.35 & 0.99 & 1.00 & 0.01 \\
  & Rec  & \textbf{0.67} & \textbf{0.35} & \textbf{0.35} & \textbf{0.35} & \textbf{1.00} \\
  & F1   & 0.27 & 0.35 & 0.51 & 0.51 & 0.02 \\
\midrule
\multirow{3}{*}{SVM\_G} 
  & Prec & 0.52 & 0.75 & 0.99 & 1.00 & 0.03 \\
  & Rec  & \textbf{0.87} & \textbf{0.75} & \textbf{0.75} & \textbf{0.75} & \textbf{1.00} \\
  & F1   & 0.46 & 0.75 & 0.85 & 0.86 & 0.06 \\
\bottomrule
\end{tabular}
\vspace{0.2em}
\end{table}

\begin{table}[t]
\centering
\caption{Test Data Confusion Matrix - *{\footnotesize TN: True Negative, FP: False Positive, FN: False Negative, TP: True Positive}}
\label{tab:test_confusion_matrix}
\begin{tabular}{llll}
\toprule
\bfseries Model & \bfseries True/Predict & \bfseries 0 & \bfseries 1 \\
\midrule
       & 0 & 110 (TN) & 20 (FP) \\
LR     & 1 & 0 (FN)   & \textbf{1 (TP)} \\
\midrule
       & 0 & 130 (TN) & 0 (FP) \\
RF     & 1 & 1 (FN)   & 0 (TP) \\
\midrule
       & 0 & 99 (TN)  & 31 (FP) \\
SVM    & 1 & 0 (FN)   & \textbf{1 (TP)} \\
\midrule
       & 0 & 43 (TN) & 87 (FP) \\
LR\_G & 1 & 0 (FN)   &  \textbf{1 (TP)} \\
\midrule
       & 0 & 45 (TN)  & 85 (FP) \\
RF\_G  & 1 & 0 (FN)   & \textbf{1 (TP)} \\
\midrule
       & 0 & 97 (TN)  & 33 (FP) \\
SVM\_G  & 1 & 0 (FN)   & \textbf{1 (TP)} \\
\bottomrule
\end{tabular}
\vspace{0.5em}
\end{table}

From Table~\ref{tab:model_test_scores}, LR shows a high macro recall (0.92), indicating good sensitivity, but low macro precision (0.52) and macro F1 score (0.50), suggesting false positives. Its micro averages for precision, recall, and F1 score (0.85) indicate balanced overall performance. SVM achieves similar macro recall (0.88) and precision (0.52) but a lower F1 score (0.46), with all micro scores at 0.76. In contrast, RF struggles with macro metrics, whose precision, recall, and F1 scores are all at 0.50 but achieves high micro scores (0.99), reflecting its bias towards the majority class (label 0).

From Table~\ref{tab:model_test_scores}, the LR\_G model shows a macro recall of 0.67, a decrease from its original data score 0.92, while macro precision drops slightly to 0.50. The macro F1 score falls further to 0.26, indicating ongoing false positive challenges. Micro precision, recall, and F1 scores are all at 0.34, showing a noticeable reduction in overall performance compared to 0.85. For SVM\_G, Table~\ref{tab:model_test_scores} reports a macro recall of 0.87, macro precision of 0.52, and macro F1 score of 0.46, nearly identical to its original-data counterpart. However, the micro scores drop slightly to 0.75, from 0.76, showing SVM\_G's ability to maintain balance across classes with minimal performance degradation. Notably, RF\_G-unlike RF trained on real data-was able to improve where the original model completely failed. Its macro recall rises to 0.67 (from 0.50), and while its macro F1 score remains modest at 0.27, this is comparable to LR\_G’s 0.26. Micro metrics for RF\_G drop to 0.35, showing reduced overall accuracy, but its performance now aligns more closely with LR\_G, suggesting that GAN-based training helped RF recover meaningful minority class detection capability. This marks a shift in behavior: from being biased toward the majority class to at least partial recognition of rare cases, even if at the cost of higher false positives.

From Table~\ref{tab:test_confusion_matrix}, all models except RF successfully predicted the single positive case in the real test set, capturing the characteristics of the critical minority class, but SVM generates more false positives (31) than LR (20), indicating its sensitivity to the minority class. On the other hand, RF underperformed, failing to detect the positive case entirely, which might be due to the RF model's nature, which aggregates multiple decision trees, where they tend to split on more common features (majority class), making them biased towards the majority class (negative samples). Consequently, RF misclassified the minor class(positive samples). 

Among the GAN-trained models, LR\_G, SVM\_G, and RF\_G all successfully identified the positive test sample. LR\_G produced the highest number of false positives (87), followed closely by RF\_G (85), while SVM\_G had 33. SVM\_G achieved a better balance by maintaining fewer false positives compared to LR\_G and RF\_G. This suggests that SVM generalizes better when trained on synthetic data, possibly due to its margin-based optimization that can adapt without overfitting to noise.

In particular, RF\_G’s ability to detect the positive case marks a significant improvement over RF trained on the original data. Although the model still produced a high number of false positives, its recall improved from 0 to 1, highlighting that the GAN-generated data introduced a meaningful signal that RF was able to learn. In summary, SVM trained on two datasets performed well on test data compared to other models.

\end{document}